\documentclass[conference]{IEEEtran}
\IEEEoverridecommandlockouts
\usepackage{cite}
\usepackage{amsmath,amssymb,amsfonts}
\usepackage{algorithmic}
\usepackage{graphicx}
\usepackage{textcomp}
\usepackage{xcolor}
\usepackage{subcaption}
\def\BibTeX{{\rm B\kern-.05em{\sc i\kern-.025em b}\kern-.08em
    T\kern-.1667em\lower.7ex\hbox{E}\kern-.125emX}}
\begin{document}

\title{Problems of representation of electrocardiograms in convolutional neural networks \\
\thanks{The study was supported by the Ministry of Education and Science of Russia (Project No. 14.Y26.31.0022).}
}
\makeatletter
\newcommand{\linebreakand}{%
  \end{@IEEEauthorhalign}
  \hfill\mbox{}\par
  \mbox{}\hfill\begin{@IEEEauthorhalign}
}
\makeatother

\author{\IEEEauthorblockN{1\textsuperscript{st} Iana Sereda}
\IEEEauthorblockA{\textit{dept. of Control Theory} \\
\textit{Nizhny Novgorod State University}\\
Nizhny Novgorod, Russia \\
sereda@itmm.unn.ru}
\and
\IEEEauthorblockN{2\textsuperscript{nd} Sergey Alekseev}
\IEEEauthorblockA{\textit{dept. of Control Theory} \\
\textit{Nizhny Novgorod State University}\\
Nizhny Novgorod, Russia \\
sergey.alekseev@itlab.unn.ru}
\and
\IEEEauthorblockN{3\textsuperscript{rd} Aleksandra Koneva}
\IEEEauthorblockA{\textit{dept. of Control Theory} \\
\textit{Nizhny Novgorod State University}\\
Nizhny Novgorod, Russia \\
aleksandra.koneva@itlab.unn.ru}
\linebreakand
\IEEEauthorblockN{4\textsuperscript{th} Alexey Khorkin}
\IEEEauthorblockA{\textit{dept. of Control Theory} \\
\textit{Nizhny Novgorod State University}\\
Nizhny Novgorod, Russia \\
alexeykhorkin12@gmail.com}
\and
\IEEEauthorblockN{5\textsuperscript{th} Grigory Osipov}
\IEEEauthorblockA{\textit{dept. of Control Theory} \\
\textit{Nizhny Novgorod State University}\\
Nizhny Novgorod, Russia \\
osipov@vmk.unn.ru}
}

\maketitle

\begin{abstract}
Using electrocardiograms as an example, we demonstrate the characteristic problems that arise when modeling one-dimensional signals containing inaccurate repeating pattern by means of standard convolutional networks. We show that these problems are systemic in nature. They are due to how convolutional networks work with composite objects, parts of which are not fixed rigidly, but have significant mobility. We also demonstrate some counterintuitive effects related to generalization in deep networks.
\end{abstract}


\section{Introduction}
Modern convolutional networks have achieved great success in pattern recognition. In recent years they have also proven themselves in sequence analysis problems\cite{piczak2015environmental}, showing some specific advantages over recurrent networks\cite{bai2018empirical}. However, the well-known issue of deep networks is the difficulty of interpreting of the resulting model. A lot of work has been done in this direction, both practical \cite{zhang2018interpretable} and theoretical \cite{dong2019geometrization}\cite{higgins2018towards}.

Convolutional networks are mainly investigated\cite{bolei2015object}\cite{zeiler2014visualizing} on tasks related to the analysis of visual scenes, i.e. photo, video. ECG diagnostic task belongs to another class of problems. It is sequence classification problem, where the signal has fuzzy cyclic structure. Convolutional networks have been successfully applied to this task \cite{acharya2017application}\cite{gadaleta2019effectiveness}\cite{7202837}. Interpretability of the model is especially important in medicine, so in this paper we focused on elucidating the details of how deep convolutional networks build a representation of signals with fuzzy cyclic structure.

In the tasks of image analysis under certain conditions a good interpretability of a deep model is achievable: the model learns deep features that are understandable to human. However, convolutional networks are known to experience fundamental difficulties in some situations. In  \cite{liu2018intriguing} it was shown that the problem arises when the network needs to determine the coordinates of some object inside the region while looking at the picture of that region. This behavior is reproducible even on a simple synthetic problem of that kind.

Doctors consider the electrocardiogram (ECG) signal as a sequence of composite objects - cardiac cycles. Each cycle usually includes several components. These objects can be located in different places of a specific ECG. This, at first glance, makes the situation similar to that considered in work\cite{liu2018intriguing}. However, for diagnosis by ECG, one does not need to know the absolute coordinates of these objects inside the ECG. Rather, one needs to know the distances between them, i.e. the problem arises of determining not absolute coordinates, but relative ones. 

There is theoretical evidence that deep learning networks work well with a periodic signal\cite{szymanski2014deep}. However, an ECG signal implies only “approximate” shear symmetry. The cardiac cycle can vary from time to time both in morphology and in duration. 

The article is organized as follows. In the first part (\ref{interp_id_section}) experiments with interpolation demonstrate the problems of representing the ECG signal by means of a deep convolutional autoencoder. These problems are stable under hyperparameters variation and do not disappear after the use of standard regularization tools. In section \ref{sec::one_cycle_interpol}, we investigate the cause of these problems with another interpolation experiment.

The section \ref{good-bad-section} is devoted to the question of whether the convolutional network exploits the inter-cyclic symmetry present in the signal.

\subsection*{ECG data}
The experiments used the ECG of patients from the LUDB dataset\cite{2018arXiv180903393K} - both those with pathologies and healthy ones. Example ECGs are shown in figure \ref{ecg_example}.
Each record ECG is 10 seconds long, with a sampling rate of 500 Hz. An ECG have three main components: the P-wave, the QRS-complex and T-wave (fig. \ref{pqrst}).

\begin{figure}[!ht]
\centering
\includegraphics[scale=0.54]{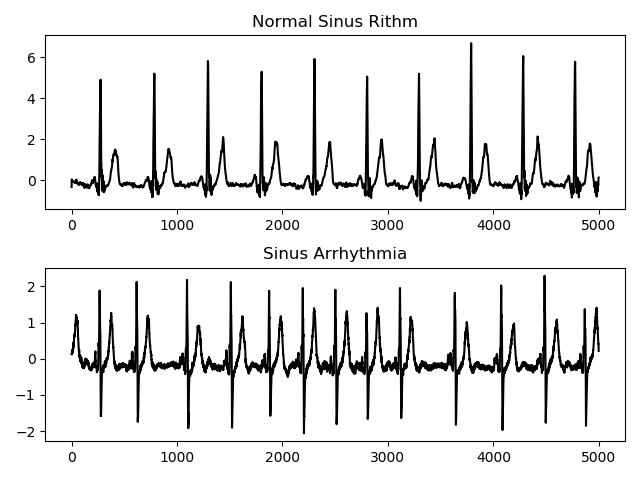}
\caption{\small{Example of data: signal of one ECG lead of a healthy patient (top) and with disease (bottom). The signal is normalized and the baseline drift is eliminated. }}
\label{ecg_example}
\end{figure}

\begin{figure}[!ht]
\centering
\includegraphics[scale=0.5]{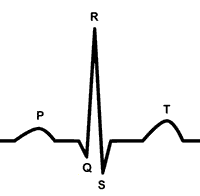}
\caption{\small{Schematic cardiac cycle.}}
\label{pqrst}
\end{figure}

The project’s code is available at: https://github.com/Namenaro/cnn-ecg

\begin{figure*}[t]
  \centering
    \begin{subfigure}{\linewidth}
    \centering
      \includegraphics[width=\linewidth]{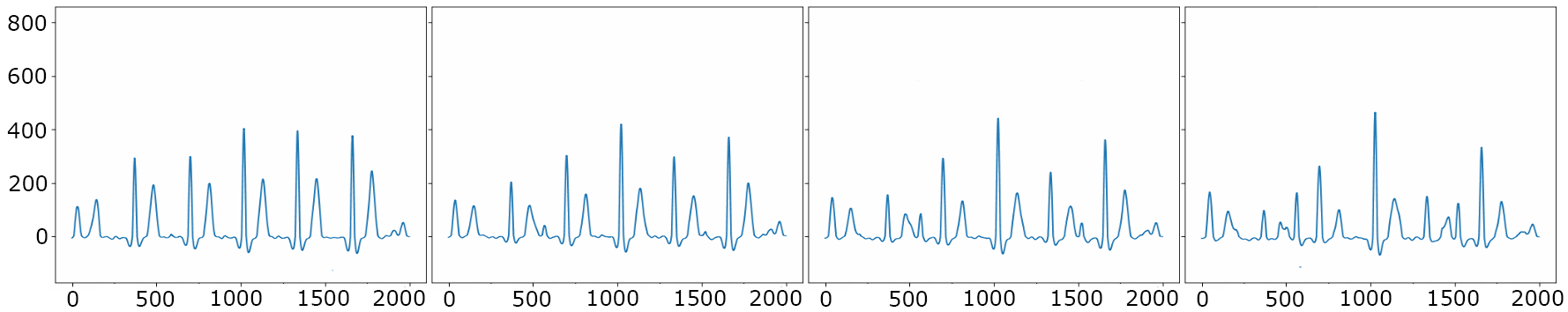}
       \hspace{4cm}{(a)}\hspace{4cm}{(b)}\hspace{4cm}{(c)}\hspace{4cm}{(d)}
    \end{subfigure}\\
    \begin{subfigure}{\linewidth}
        \centering
\includegraphics[width=\linewidth]{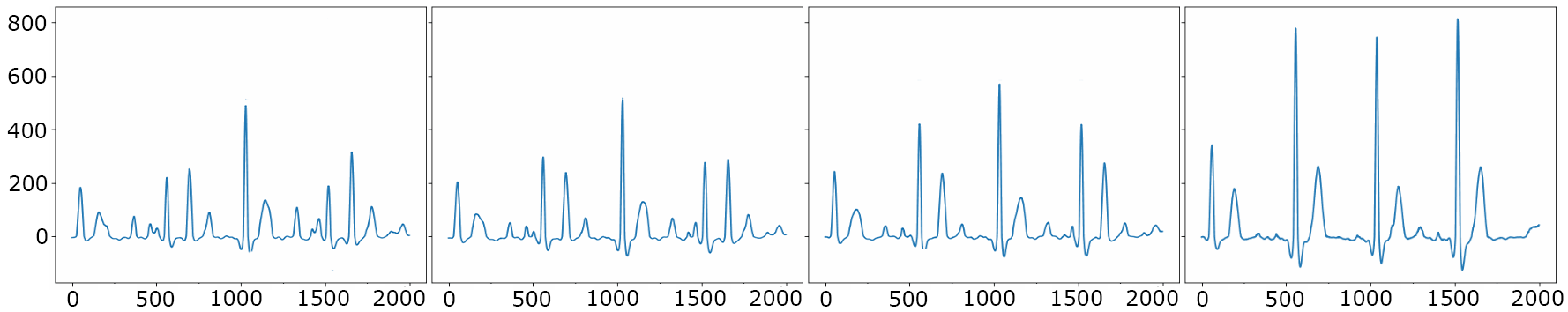}
        \hspace{4cm}{(e)}\hspace{4cm}{(f)}\hspace{4cm}{(g)}\hspace{4cm}{(h)}
    \end{subfigure}\\
  \caption{Example of generated images by the encoder from eight points of a straight line between the representation of the first ECG (a) and the second ECG (h) in the latent space of the autoencoder. The points of the straight line are taken  sequentially from the first ECG to the second one. Thus, images (b)-(d) are generated by the decoder from intermediate values. It can be seen that not all images correspond to the ECG signal.}
  \label{full_interpolation_1d}
\end{figure*}

\section{Interpolation between ECGs} \label{interpolation_section}
One of the standard means of assessing the quality of representation is interpolation between points in the dataset by a generative model. If the structure of the model is chosen correctly, then the resulting representation tends to be disentangled (and therefore interpretable by a human) and satisfies the manifold hypothesis. 

It was empirically shown in \cite{shao2018riemannian}, that in the case when the deep network is able to successfully parameterize the manifold in the data space, the curvature of the manifold is small. As a result of this, straight segments in latent space correspond quite closely to geodesics on that manifold. This observation provides a tool for exploring the learned features of a deep generative model.

\subsection{Interpolation by means of 1D-convolutional network} \label{interp_id_section}

A simple convolutional autoencoder was built and trained on 4-second ECG signals of lead V6. The encoder model consists of five blocks (table \ref{base_table}): the first four blocks include one-dimensional convolutions with ReLU activation function, batch normalization along the channels axis and max pooling with pooling window 2 and the last block is a fully connected layer with dimensionality of the output space 30. The structure of the decoder is symmetrical. 

\begin{table}[!ht]
\caption{Structure of encoder. Regularisation layers are not shown.}
    \label{base_table}
    \centering
\begin{center}
\begin{tabular}{|c|c|c|}
\hline
\textbf{Type} & \textbf{Parameters} & \textbf{Input size} \\ \hline
conv          & 100x1/30            & 512x1               \\ \hline
pool          & 2                   & 512x30              \\ \hline
conv          & 100x1/15            & 256x30              \\ \hline
pool          & 2                   & 256x15              \\ \hline
conv          & 30x1/15             & 128x15              \\ \hline
pool          & 2                   & 128x15              \\ \hline
conv          & 20x1/5              & 64x15               \\ \hline
pool          & 2                   & 64x5                \\ \hline
dense         & 30                  & 32x5                \\ \hline
\end{tabular}
\end{center}
\end{table}
The trained model showed good reconstruction on the test and training part of the patients: all significant peaks were reconstructed recognizably, the model eliminated high-frequency noise from the signal.

The algorithm of the interpolation experiment was as follows:

\begin{enumerate}
\item two ECGs are randomly selected from the data set
\item using the encoder, the corresponding coordinates in the latent space are calculated
\item a straight line is drawn between these two points
\item using a decoder, points of this segment are decoded into the data space
\end{enumerate}

 From the assumption that straight lines in the latent space correspond quite closely to some paths on the learned data manifolds, the decoded points from this line should also be decoded into ECGs.

Figure \ref{full_interpolation_1d} shows these decoded states. Signals from a straight line in the latent space at some point in time (pretty fast) cease to be ECGs (fig. \ref{full_interpolation_1d}d-f). When considering the transition from one ECG to the other, it was found that the segments gradually disappears in one place and gradually "grows” in another. In other words, the number of important peaks was changing during the interpolation and the  periodical structure (repeating cardiac cycle) had disappeared in some moment.

On the assumption that this behavior is influenced by the repeating structure in the signal, similar experiments were conducted, but with a single centered cardiac cycle.

\subsection{Cropped signal interpolation} \label{sec::one_cycle_interpol}
For this experiment, the data was generated as follows: all R-peaks were found on the ECG, the found R peaks and 1-second areas around them were used as new data (This interval was chosen because it is the average length of the RR interval of a healthy person). 

In this experiment, there is no repeating complex structure in the signal, because only one cardiac cycle is involved. The following behavior was registered: when the segment (a wave, a complex of waves) of the first ECG and the corresponding segment of the second ECG are placed at a distance smaller than their length, than the network “moves” the wave from one position to another. That is, at every moment the wave exists, but its center of mass drifts from position A to position B (see fig. \ref{interpolation_1d}\subref{move}).

If there is a large distance in time between the corresponding complexes of the two ECGs, then during interpolation the complex gradually disappears in one place and “grows” in another, as in 4-second ECG signal (fig. \ref{interpolation_1d}\subref{retract}).

\begin{figure}[ht!]
     \begin{subfigure}[b]{0.5\textwidth}
        \centering
         \includegraphics[scale=0.25]{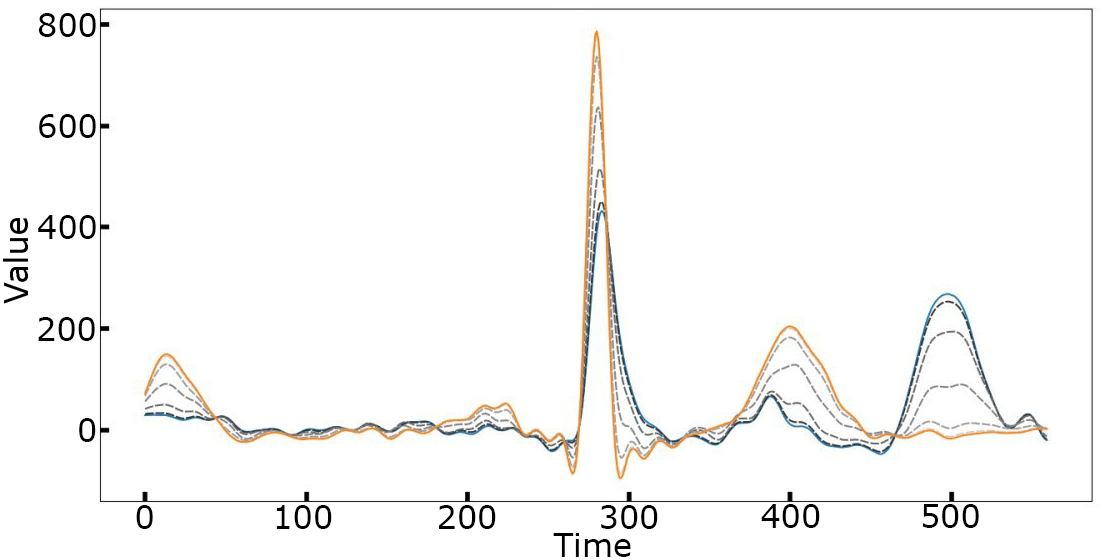}
         \caption{\small{T-wave retracts in one position and appears in another position}}
         \label{retract}
     \end{subfigure}
     \hfill
     \begin{subfigure}[b]{0.5\textwidth}
     \centering
         \includegraphics[scale=0.25]{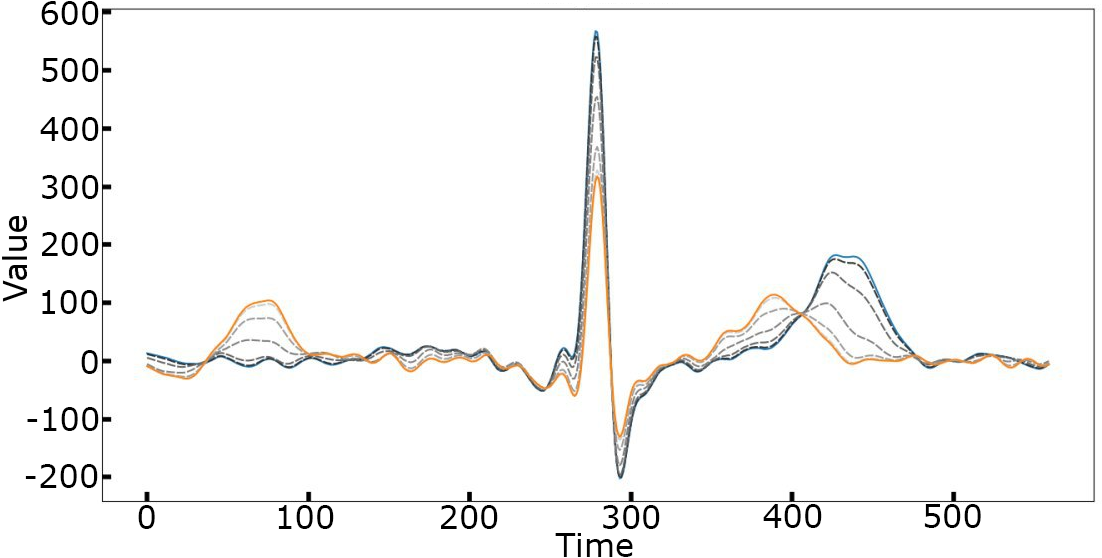}
         \caption{\small{T-wave "moves" from one position to another}}
         \label{move}
     \end{subfigure}
     \caption{\small{Interpolation between two signals by means of 1d-convolutional network. Shades of gray correspond to the interpolation steps. Signals between which interpolation is performed are highlighted in orange and blue. The rightmost peak is the T-wave. QRS-complex is in the middle. QRS-complex does not need to move, it only needs to change the morphology and it changes it smoothly. The T-wave needs to change the morphology and to move, but this is not what happens: it disappears in one place and appears in another, there is no movement.}}
     \label{interpolation_1d}
\end{figure}

We experimentally investigated whether the variation of hyperparameters of convolutional architecture affects the described effect. It turned out that no: neither the size of the convolutional kernels, nor their number, nor the compression ratio had a significant effect.

One-cycle interpolation result (shown in fig. \ref{interpolation_1d}) indicates that the problem of interpolation between two long ECGs is not in the presence of periodicity, but in the fact that the components of the cardiac cycle are not located on rigidly fixed places inside the cycle. Indeed, the distance between the R-peak and T-peak can vary greatly depending on the pathology of the cardiovascular system. A small variation of this distance of the neural network is not terrible (we see it on \ref{move}). However, when the distance between the center of mass of the given wave on two ECGs becomes comparable to the size of the time interval containing the wave, the network interpolates incorrectly.

The described feature of the network’s behavior on one cycle determines its behavior on the entire signal with multiple cycles: at some stage the simulated cyclic structure will disappear with a high probability. That probability depends on the variability of the signal structure: in case of ECG, the normal heart rate is 60-100 beats per minute, therefore two different 4-seconds ECGs may have a different number of complexes with different distances between them. Fig.\ref{ecg_example}) shows how some pathologies can affect the regularity of distance between the complexes. 

Therefore, when interpolating between two 4-second electrocardiograms of healthy or unhealthy persons, it is impossible to align the ECG (by shift). Such alignment would be possible to some extent only in a homogeneous group of healthy patients - and only on relatively short duration of the signals.

\section{One periodicity-dependent effect in autoencoders} \label{good-bad-section}
The experiments of the previous section showed that the hyperparameters of the network do not affect the qualitative properties of interpolation. However, they affect the capacity of the network and, therefore, they should affect its ability to generalize and to overfit. Here we found an unusual effect, to which the rest of the article is devoted.

It was shown in \cite{ulyanov2018deep} that if the network structure “fits well” with the data, then after retraining it on a very small amount of data, it shows a good generative ability in a special task: to restore those regions of the input image that did not contribute to the error functional.

We modified the auto-encoder error functional: now the usual mean squared error in some regions of the signal is multiplied by zero. Thus, these regions no longer contribute to the calculation of the error gradient. In other words, in these regions the network can output anything, so we will call them "unpunished". It turned out that different network architectures react to this modification in different ways.

There is a dependence between the reconstruction quality and the degree of signal compression: the network with a small size of the pooling in the middle layers reconstructs ECG well both inside the "unpunished" area and outside. But with increase the compression ratio, the error inside the "unpunished" area increases without affecting the error outside. The differences are shown in the fig.\ref{bad_good_example}.

\begin{figure}[!ht]
\centering
\includegraphics[scale=0.43]{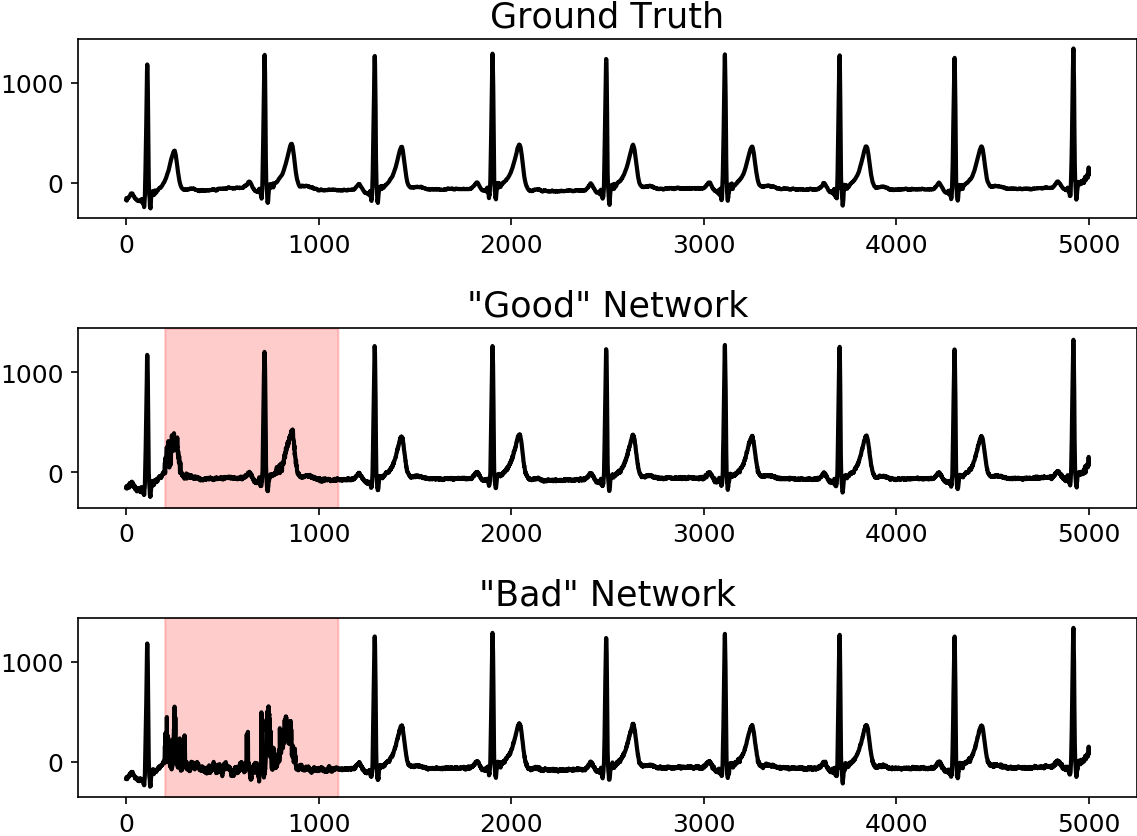}
\caption{\small{Example of ECG processing by "good" and "bad" networks. The real ECG is shown at the top. Two other images are outputs of two different convolutional autoencoders. Red area is unpunished (doesn't affect error gradients). This example shows the difference between two convolutional autoencoders with different pooling rate, which appears only within the unpunished area. It's clearly seen that “bad” and “good" networks behave identically outside the red region, but only "good" network preserves the signal structure inside the unpunished area.}}
\label{bad_good_example}
\end{figure}

\begin{figure}[!ht]
\centering
\includegraphics[scale=0.43]{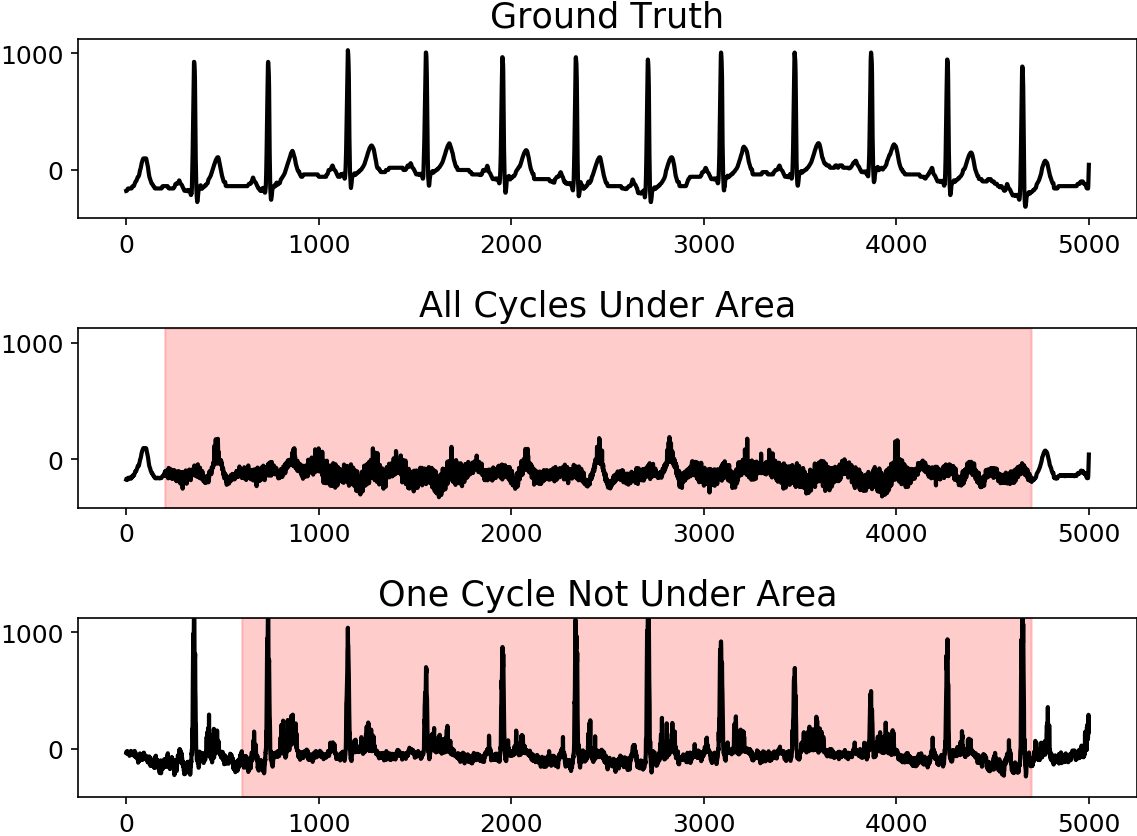}
\caption{\small{Example of ECG processing by "good" network with big size of "unpunished" area. The real ECG is shown at the top. Middle image shows an output of the autoencoder with "unpunished" area, which covered all cardiac cycles. Bottom image shows an output of the autoencoder with one cardiac cycle out of "unpunished" area. Red area is unpunished (doesn't affect error gradients). This example shows the influence of area size.}}
\label{big_mask_example}
\end{figure}

\begin{figure*}[ht!]
     \begin{subfigure}[b]{0.5\textwidth}
     \centering
         \includegraphics[scale=0.32]{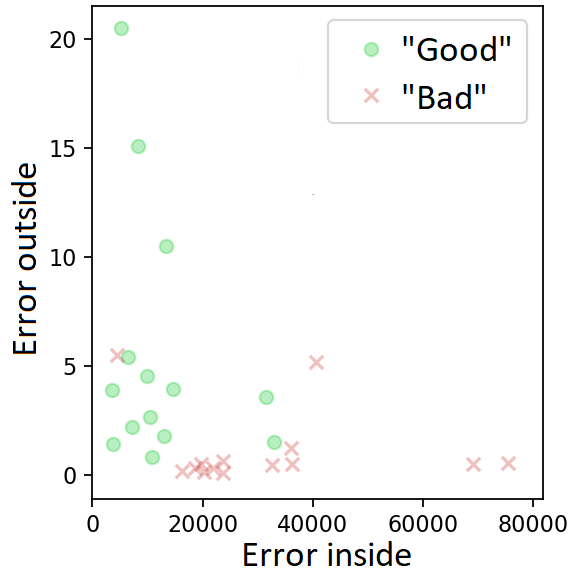}
         \caption{\small{On the ECG signal}}
         \label{scatter_ECG}
     \end{subfigure}
     \hfill
     \begin{subfigure}[b]{0.5\textwidth}
     \centering
         \includegraphics[scale=0.32]{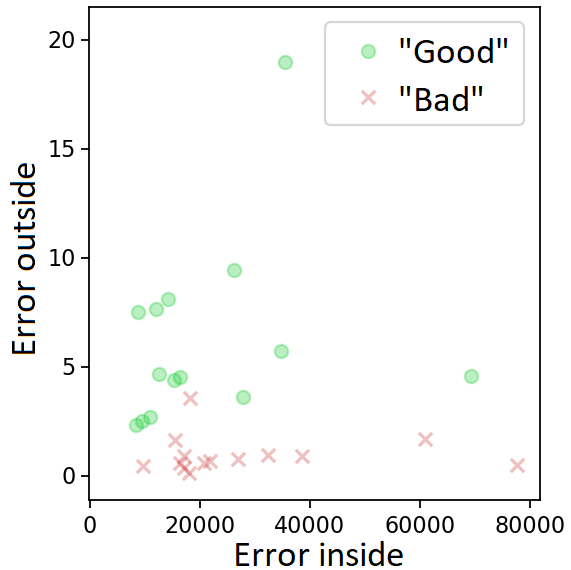}
         \caption{\small{On a non-periodic signal}}
         \label{scatter_unperiodic}
     \end{subfigure}
     \caption{\small{A scatter plot of errors inside the "unpunished" area and outside. Each point corresponds to one ECG. On the ECG signals (a), all points of a "good" network are grouped in a dense cloud with a low error inside the "unpanished" area. On a non-periodic signal (b) the error inside the "unpanished" area is almost as high as outside of it. Green cloud in (a) concentrates in the left while in (b) it is not.}}
     \label{unperiodic}
\end{figure*}

To test the dependence of this effect on the cyclic pattern of the signal, we repeated this experiment with a non-periodic signal. We sampled the trajectories of different Gaussian processes (with different kernels). The duration of each trajectory coincided with the duration of the ECG. The resulting trajectories were normalized in such a way that the average and variance for each of them coincided with the corresponding values for the ECG. Then "good" and "bad" networks were trained on these trajectories using the same protocol as for ECGs, described above. Result shown in the figures \ref{unperiodic} and \ref{unperiodic_example}. 

The behavior of a "bad" network remains unchanged: it is the same on ECG and a non-periodic signal. But for a "good" network the picture now is different: the error inside the "unpunished" area on a cyclic signal is systematically lower. There are no significant changes in the punishable area.  (fig.\ref{unperiodic}\subref{scatter_unperiodic}).

\begin{figure}[!ht]
\centering
\includegraphics[scale=0.21]{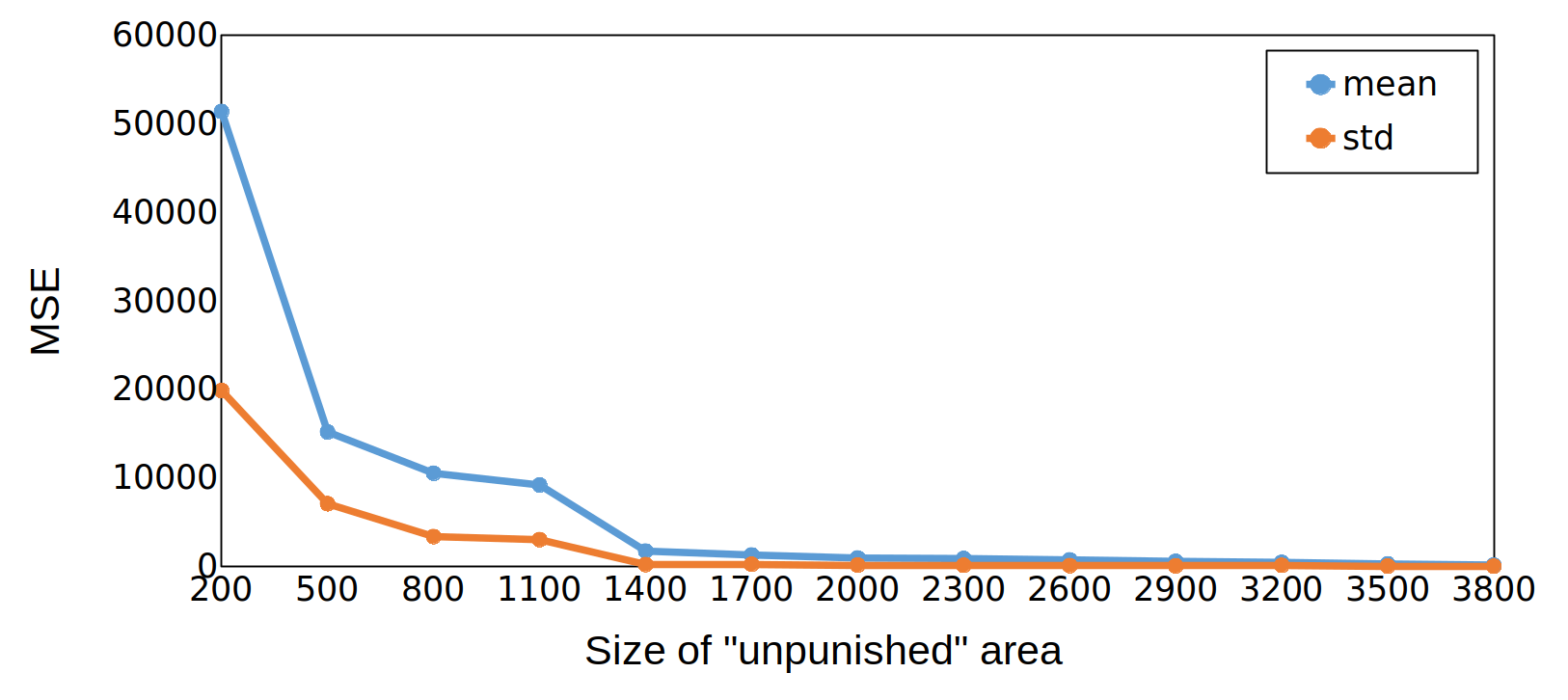}
\caption{\small{The figure shows how the error of a good network inside an unpunished area changes when the size of this area changes. The error drop curve of the network becomes much less steep from the moment when the network begins to “see” one full cycle of the electrocardiogram from under the "unpunished" area. 10 experiments were performed for each area size, then their results were averaged. Averaged over these experiments, the mean and std for each "unpunished" area size are shown in the figure. }}
\label{size_in}
\end{figure}

Simply put, a bad network is bad everywhere - both on a random signal and on ECG. A good network is bad only on a random signal. On a signal containing a repeating pattern, good network demonstrates an unexpected property - it restores even those fragments of the signal that it was not "asked" to restore.

The described effect in good network depends nonlinearly on the size of the unpunished region.  When "unpunished" area covers all cardiac cycles, reconstruction error is large, but when  at least one full signal period appears outside that region, reconstructed signal becomes much better(fig. \ref{big_mask_example}). 

With further reducing "unpunished" area, reconstruction error drops much less, even when new peaks appear in "observed" area (fig. \ref{size_in}). This behavior is also stable and does not depend on a specific electrocardiogram.

The reconstruction error outside the unpunished area predictably increases slightly as this area itself decreases.

\begin{figure}[!ht]
\centering
\includegraphics[scale=0.43]{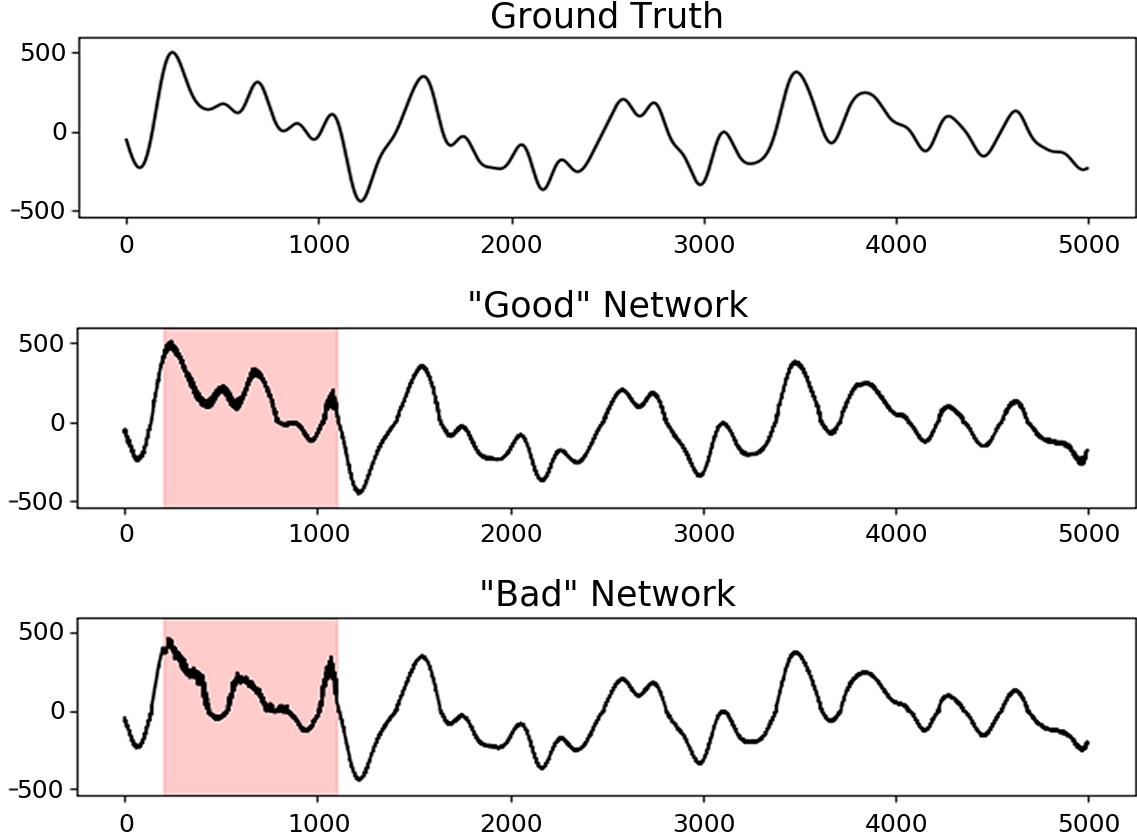}
\caption{\small{Example processing of non-periodic signal by ”good” and ”bad” networks - the networks are of the same architectures as in \ref{bad_good_example},\ref{big_mask_example}. The  real gaussian trajectory  is  shown  at  the  top.  Two  other  images  are  outputs of two different convolutional autoencoders. Red area is unpunished (doesn’t  affect  error  gradients). This example shows a lesser manifestation of the described effect on non-periodic signal.}}
\label{unperiodic_example}
\end{figure}

\section{Conclusion}\label{sec::conclusion}
Experiments in section \ref{interpolation_section} have shown that basic convolutional networks are poorly suited to the problem of modeling of signals with fuzzy cyclic structure.  That does not mean that their representation cannot be used in the transfer learning techniques, but almost certainly means it will not be interpretable.

Sec. \ref{interp_id_section} provides description of what damages interpretability when modeling a fuzzy cyclic signal by means of a convolutional network. Wide class of ECG was used as an example.

Section \ref{sec::one_cycle_interpol} empirically shows the cause of this problem. Its source lies in the mechanism of how basic convolutional networks model scenes consisting of moving components. This is shown by the example of a moving T-wave of the cardiac cycle. 

The section \ref{good-bad-section} describes an interesting interplay between hyperparameters of the network and the way of how it reconstructs ECG signal. It turned out that even if one does not punish the network practically over the entire length of the signal, it still reconstructs it with good quality under three conditions:

\begin{enumerate}
\item the signal must contain repeating pattern
\item at least one period must get into that very small punishable area
\item the compression factor does not have to be very large (but can still be significant)
\end{enumerate}

The exact nature of this effect still awaits a theoretical explanation.

\section{Discussion}\label{sec::discussion}
It is known that for the diagnosis of certain diseases by ECG, a small number of features is enough. For example, for a myocardial infarction, this number can be reduced to seven \cite{savostin2019using}. This indirectly indicates the presence of a low-dimensional structure in ECGs. But our experiments have demonstrated that basic convolutional autoencoders fail to learn interpretable parametrisation of the ECG-s manifold in the space of all signals of predefined length. Experiments indicate that common regularization techniques (like batch normalization\cite{ioffe2015batch}) do not help to eliminate the negative effects from sec.\ref{interpolation_section}. Probably, some other regularization must be developed for this situation. 

The most direct solution to the problem is, probably, to use more complicated convolutional architectures. In \cite{moskalenko2019deep} it was demonstrated, that U-Net-like architecture is capable to achieve high results in the task of ECG segmentation. In this work we investigated only the problems of the basic convolutional networks - i.e. networks of the  “linear” multi-layer convolutional structure without complications like in \cite{moskalenko2019deep}. But it was shown that even such basic architectures show a satisfactory result on the same ECG segmentation task\cite{sereda2019ecg}, and very good results - in ECG-based biometric authentication systems\cite{li2020toward}. Segmentation results on healthy patients even in such networks were high, although the result was somewhat worse on unhealthy ones. As we have shown here, that result is combined with poor ECG representation, which makes it relevant to search for a way to improve the representation in convolutional networks in general.

In ECG case there is a repeating nonrigid structure (cardiac cycle), usually consisting of three moving objects: the P-wave, the QRS complex and the T-wave. The word "usually" here means that with some pathologies, some components can degenerate. From patient to patient and from disease to disease, these objects can change their morphology, size and location within the cycle. However, within the framework of one pathology, the set of these components should be unchanged in all ECGs containing this pathology. A specific pathology of the cardiovascular system should give us a specific invariant that persists during transformations. But how to make the network parameterize this invariant?

When interpolating between two ECGs, each of which contains all three components of the cardiac cycle, none of these components should disappear or clone. The number of waves must be an invariant that is maintained during the entire interpolation. Per-pixel changes occurring between interpolation steps are changes that must preserve the number of waves and their order. This article shows that this is precisely what does not happen. In some cases, the wave disappears in one place, and appears in another, instead of drifting smoothly.

If one introduces a condition that makes every important wave drift during interpolation, it would probably solve a significant part of the problems of representing signals with fuzzy cyclic structure. The use of optimal transportation theory looks particularly promising in that sense. It could probably be used to enforce the interpolation trajectory of the network to coincide with the solution of the stochastic optimal control problem of moving the "masses" of each wave to a new location. We leave this as a direction for the future work.

\bibliographystyle{IEEEtran}
\bibliography{references}

\end{document}